\documentclass[a4paper, 10pt, conference]{ieeeconf}      

\title{City-scale Road Extraction from Satellite Imagery} 

\author{Adam Van Etten \\
CosmiQ Works, In-Q-Tel\\
{\tt\small avanetten@iqt.org}
}

\usepackage{cite}
\usepackage{times}
\usepackage{epsfig}
\usepackage{graphicx}
\usepackage{amsmath}
\usepackage{amssymb}

\usepackage{booktabs}       
\usepackage{threeparttable}  
\usepackage{paralist}
\usepackage{verbatim}
\usepackage{comment}
\usepackage{amsmath}
\usepackage{amssymb}
\usepackage{url}     
\usepackage{algorithm}
\usepackage{algpseudocode}

\usepackage{subfig}

\begin{document}

\maketitle
\thispagestyle{empty}
\pagestyle{empty}

\begin{abstract}

Automated road network extraction from remote sensing imagery remains a significant challenge despite its importance in a broad array of applications.
To this end, we leverage recent open source advances and the high quality SpaceNet dataset to explore road network extraction at scale, an approach we call City-scale Road Extraction from Satellite Imagery (CRESI).    Specifically, we create an algorithm to extract road networks directly from imagery over city-scale regions, which can subsequently be used for routing purposes.  We quantify the performance of our algorithm with the APLS and TOPO graph-theoretic metrics over a diverse 608 square kilometer test area covering four cities.  We find an aggregate score of APLS = 0.73, and a TOPO score of 0.58 (a significant improvement over existing methods).  Inference speed is $\geq$160 square kilometers per hour on modest hardware.  Finally, we demonstrate that one can use the extracted road network for any number of applications, such as optimized routing.  

\end{abstract}

\section{Introduction}\label{sec_intro}

Determining optimal routing paths in near real-time is at the heart of many humanitarian, civil, and commercial challenges. In the humanitarian realm, for example, traveling to disaster stricken areas can be problematic for relief organizations, particularly if flooding has destroyed bridges or submerged thoroughfares. Autonomous vehicle navigation is one of many examples on the commercial front, as self-driving cars rely heavily upon highly accurate road maps. Existing data collection methods such as manual road labeling or aggregation of mobile GPS tracks are currently insufficient to properly capture either underserved regions (due to infrequent data collection), or the dynamic changes inherent to road networks in rapidly changing environments. We believe the frequent revisits of satellite imaging constellations may accelerate existing efforts to quickly update road network and routing information. Yet while satellites in theory provide an optimal method to rapidly obtain relevant updates, most existing computer vision research methods for extracting information from satellite imagery are neither fully automated, nor able to extract routing information from imaging pixels.

Current approaches to road labeling are often manually intensive. Classical computer vision techniques such as thresholding and edge detectors are often very helpful in identifying rough outlines of roads, though usually require manual inspection and validation. In the commercial realm, projects such as Bing Maps and Google Maps have been very successful in developing road networks from overhead imagery, though such processes are still labor intensive, and proprietary.

\begin{figure}
  \centering
     \includegraphics[width=0.95\linewidth]{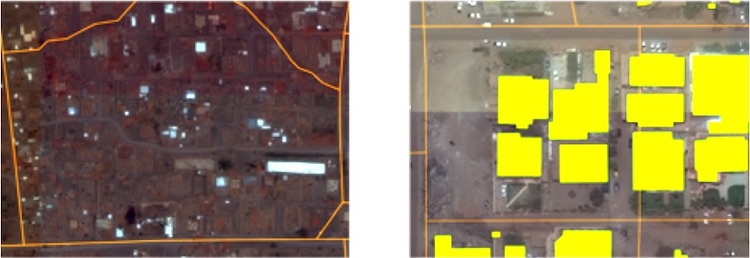}
  \caption{Potential issues with OSM data.  Left: OSM roads (orange) overlaid on Khartoum imagery; the road traveling left to right across the image is missed.  Right: OSM road labels (orange) and SpaceNet building footprints (yellow); in some cases road labels are misaligned and pass through buildings.}
   \label{fig:osm_goof}
   \vspace{-6pt}
\end{figure}

On the open source side, OpenStreetMap (OSM) \cite{OpenStreetMap} is an extensive data set built and curated by a community of mappers. It primarily consists of streets, but contains some building, point of interest, and geographic labels as well. The OSM community developed a robust schema for roads, including the type of road (e.g., residential, highway, etc), as well as other metadata (e.g., speed limit).
For many regions of the world, OSM road networks are remarkably complete, though in developing nations the networks are often poor. Regardless of region, OSM labels are at times outdated due to dynamic road networks, or poorly registered with overhead imagery (i.e., labels are offset from the coordinate system of the imagery), see Figure \ref{fig:osm_goof}.

In dynamic scenarios (such as natural disasters) where timely high quality road network revisions are crucial, the manual and semi-automated techniques discussed above often fail to provide updates on the requisite sufficient timescale. 
A fully automated approach to road network extraction therefore warrants investigation, and is explored in the following sections.

\subsection{Existing Approaches and Datasets}

Existing publicly available labeled overhead or satellite imagery datasets tend to be relatively small, or labeled with lower fidelity than desired for foundational mapping.  For example, the ISPRS semantic labeling benchmark \cite{isprs_sem}
dataset contains high quality 2D semantic labels over two cities in Germany and covers a compact area of 4.8 km$^2$; 
imagery is obtained via an aerial platform and is 3 or 4 channel and  5-10 cm in resolution.  The TorontoCity Dataset \cite{torontocity} contains high resolution 5-10 cm aerial 4-channel imagery, and $\sim700$ km$^2$ of coverage; building and roads are labeled at high fidelity (among other items), but the data has yet to be publicly released.  The Massachusetts Roads Dataset \cite{MnihThesis} contains 3-channel imagery at 1 meter resolution, and $2600$ km$^2$ of coverage; the imagery and labels are publicly available, though labels are scraped from OpenStreetMap and not independently collected or validated. 
The large dataset size, higher resolution (0.3 meter resolution), and hand-labeled and quality controlled labels of SpaceNet \cite{spacenet} provide a significant enhancement over current datasets and provide an opportunity for algorithm improvement. 

Extracting road pixels in small image chips from aerial imagery has a rich history (e.g. 
\cite{zhang2017}, 
\cite{mattyus16} 
\cite{wang2016}, 
\cite{zhang2017}, 
\cite{sironi14}, 
\cite{mnihroads}). 
These algorithms generally use a segmentation + post-processing approach combined with lower resolution imagery (resolution $\geq 1$ meter), and OpenStreetMap labels. 

Extracting road networks directly has also been attempted by a number of studies.  
\cite{stoica04} 
attempted road extraction via a Gibbs point process, while 
\cite{wegner13} 
showed some success with road network extraction with a conditional random field model. 
\cite{chai13} 
used junction-point processes to recover line networks in both roads and retinal images
\cite{turet13}  
extracted road networks by representing image data as a graph of potential paths.
\cite{mattyus15} 
extracted road centerlines and widths via OSM and a Markov random field process.
\cite{mosinska18} 
 used a topology-aware loss function to extract road networks from aerial features as well as cell membranes in microscopy.  

Of greatest interest for this work are recent papers by 
\cite{deeproadmapper} 
and \cite{roadtracer}.  \cite{deeproadmapper} used a segmentation followed by $A^{*}$ search, applied to the not-yet-released TorontoCity Dataset.  
The RoadTracer \cite{roadtracer} paper 
utilized an interesting approach that used OSM labels to directly extract road networks from imagery without intermediate steps such as segmentation.  While this approach is compelling, according to the authors it ``struggled in areas where roads were close together'' \cite{fbastani_roads} and underperforms other  techniques such as segmentation+post-processing
when applied to higher resolution 
data with dense labels.   Given that \cite{roadtracer} noted superior performance to \cite{deeproadmapper}, we compare our results to the RoadTracer paper.  

\subsection{DataSet}

We use the SpaceNet 3 dataset, comprised of 30 cm WorldView3 DigitalGlobe satellite imagery and attendant road centerline labels.  Imagery covers 3000 square kilometers, and over 8000 km of roads are labeled \cite{spacenet}.  Training images and labels are tiled into $1300 \times 1300$ pixel ($\approx400 \, \rm{m}$) chips.  

\section{Narrow-Field Baseline Algorithm}\label{sec:algo}

\begin{figure}[]
  \centering
     \includegraphics[width=0.95\linewidth]{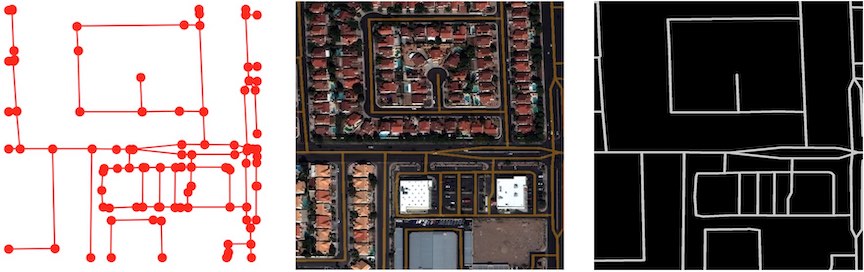}
  \caption{Left: SpaceNet GeoJSON label.  Middle: Image with road network overlaid in orange.  Right. Segmentation mask of road centerlines.}
    \label{fig:baseline_train}
    \vspace{-6pt}
\end{figure}

We initially train and test an algorithm on the SpaceNet 
chips.
Our approach is to combine recent work in satellite imagery semantic segmentation with improved post-processing techniques for road vector simplification.  
We begin with inferring a segmentation mask using convolutional neural networks (CNNs).  We assume a road centerline halfwidth of 2m and create training masks using the raw imagery and SpaceNet geoJSON road labels (see Figure \ref{fig:baseline_train}).  We create a binary road mask for each of the 2780 SpaceNet training image chips. To train our model we cast the training masks into a 2-layer stack consisting of source (road) and background.

\begin{figure}[]
  \centering
     \includegraphics[width=0.95\linewidth]{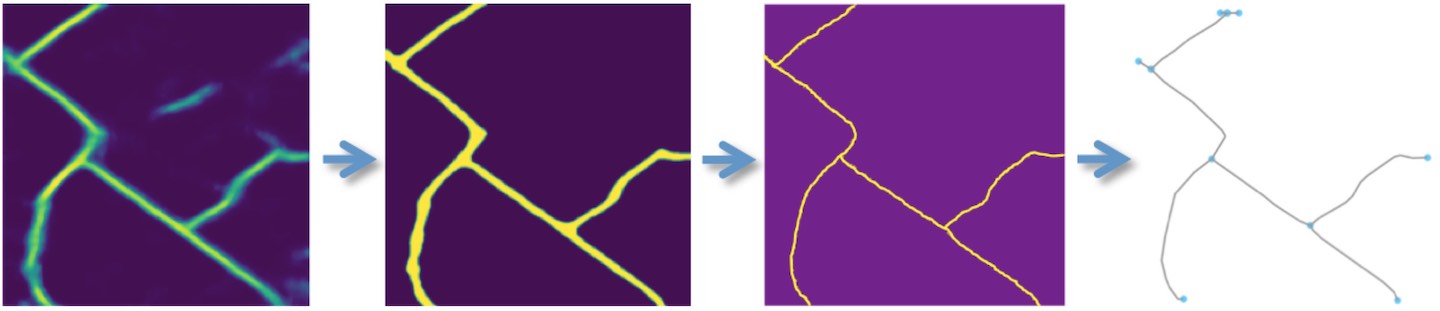}
  \caption{Initial baseline algorithm.  Left: Using road masks, we train CNN segmentation algorithms (such as PSPNet \cite{pspnet} or U-Net \cite{unet}) to infer road masks from SpaceNet imagery.  Left center: These outputs masks are then refined using standard techniques: thresholding, opening, closing, and smoothing.  Right center: A skeleton is created from this refined mask (e.g.: sckit-image skeletonize \cite{skimage}.
  Right: This skeleton is subsequently rendered into a graph structure with the {\it sknw} package \cite{sknw}.  
}
    \label{fig:baseline}
    \vspace{-6pt}
\end{figure}

We train a segmentation model inspired by the winning SpaceNet 3 algorithm \cite{albu}, and use
a ResNet34 \cite{resnet} encoder with a U-Net \cite{unet} inspired decoder.  We include skip connections every layer of the network, with an Adam optimizer and a custom loss function of:
\begin{equation}
\label{eqn:c}
 \mathcal{L} =0.8 \times {\rm BCE} \, + \,0.2 \times (1 - {\rm Dice})
\end{equation}
where `BCE' is binary cross entropy, and `Dice' is the Dice coefficient.
3-band RGB SpaceNet training data is split into four folds (1 fold is a unique combination of 75\% train + 25\% validation) and training occurs for 30 epochs.  At inference time the 4 models from the 4 folds are merged by mean to give the final road mask prediction.
These road mask predictions are then refined into road vectors via the process described in Figure \ref{fig:baseline}.

\begin{figure}[]
  \centering
     \includegraphics[width=0.95\linewidth]{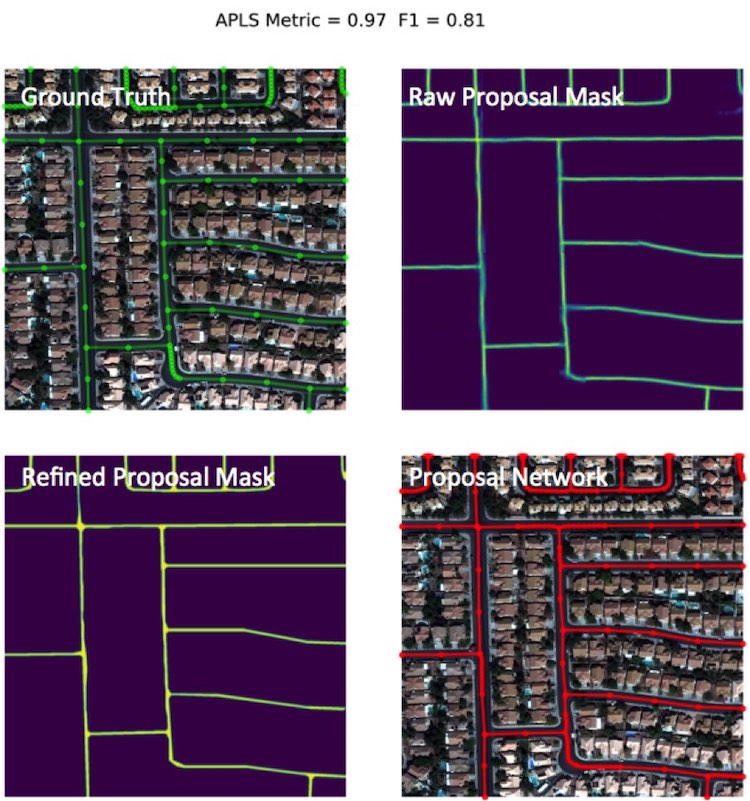}
  \caption{Sample output of baseline algorithm applied to SpaceNet test data.}
    \label{fig:baseline_ex}
   \vspace{-6pt}
\end{figure}

We also attempt to close small gaps and remove spurious connections not already corrected by the opening and closing procedures.  As such, we remove disconnected subgraphs with an integrated path length of less than 80 meters.  We also follow \cite{albu} and remove terminal vertices if it lies on an edge less than 10 pixels in length, and connect terminal vertices if the distance to the nearest non-connected node is less than 20 pixels.  The final narrow-field baseline algorithm consists of the steps detailed in Table \ref{tab:algo}; an example result is displayed in Figure \ref{fig:baseline_ex}.

\begin{table}[]
  \caption{Baseline inference algorithm}
  \label{tab:algo}
  \small
  \centering
   \begin{tabular}{ll}
    \toprule
     Step & Description \\
    \toprule
 	1 & Apply the 4 trained segmentation models to test data \\
	2 & Merge these 4 predictions into a total road mask \\
	3 & Clean road mask with opening, closing, smoothing \\ 
	4 & Skeletonize road mask \\ 
	5 & Extract graph from skeleton \\ 
	6 & Remove spurious edges and close small gaps in graph \\
    \bottomrule
  \end{tabular}
\end{table}

\section{Evaluation Metrics}\label{sec:metrics}

Historically, pixel-based metrics (such as F1 score) have been used to assess the quality of road proposals, though such metrics are suboptimal for a number of reasons (see \cite{spacenet} for further discussion).  Accordingly, we use the graph-theoretic  Average Path Length Similarity (APLS) and map topology (TOPO) \cite{topo_metric} metrics that were created to measure similarity between ground truth and proposal road graphs. 

\subsection{APLS Metric}\label{sec:apls}


\begin{figure}
  \centering
     \includegraphics[width=0.95\linewidth]{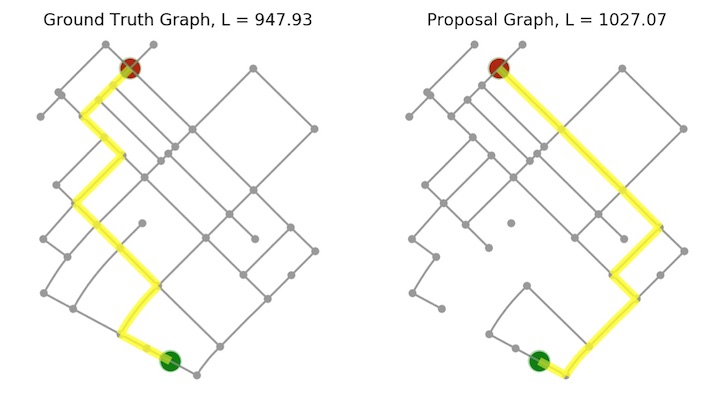}
  \caption{\textbf{APLS metric.} APLS compares path length difference between sample ground truth and proposal graphs. Left: Shortest path between source (green) and target (red) node in the ground truth graph is shown in yellow, with a path length of $\approx948$ meters. Right: Shortest path between source and target node in the proposal graph with 30 edges removed, with a path length of $\approx1027$ meters; this difference in length forms the basis for the metric  \cite{spacenet}.
  }
  \label{fig:apls_fig0}
\vspace{-6pt}
\end{figure}

To measure the difference between ground truth and proposal graphs,  the 
 APLS \cite{spacenet}
 metric sums the differences in optimal path lengths between nodes in the ground truth graph G and the proposal graph G',
 with missing paths in the graph assigned a score of 0.
 The APLS metric scales from 0 (poor) to 1 (perfect).
Missing nodes of high centrality will be penalized much more heavily by the APLS metric than missing nodes of low centrality. 
For the SpaceNet 3 challenge midpoints were injected along each edge of the graph, and routes computed between all nodes in the graph.  For the large graphs we consider here this approach is computationally intractable, so in this paper we refrain from injecting midpoints along edges, and select 500 random control nodes.
This approach yields an APLS score $\approx3-6\%$ lower than the greedy approach since midpoints along the graph edges tend to increase the total score.  
In essence, APLS measures the optimal routes between nodes of interest, so in our effort to determine optimal routes directly from imagery, this metric provides a good measure of success.

\subsection{TOPO Metric}

The TOPO metric \cite{topo_metric} is an alternative metric for computing road graph similarity.  TOPO compares the nodes that can be reached within a small local vicinity of a number of seed nodes, categorizing proposal nodes as true positives, false positives, or false negatives depending on whether they fall within a buffer region (referred t to as the ``hole size''). By design, this metric evaluates local subgraphs in a small subregion ($\sim 300$ meters in extent), and relies upon physical geometry.  Connections between greatly disparate points ($>300$ meters apart) are not measured.

\section{Comparison with OSM}\label{sec:osm}

OpenStreetMap (OSM) is a great crowd-sourced resource curated by a community of volunteers, and consists primarily of hand-drawn road labels.
Though OSM is a great resource, it is incomplete in many areas (see Figure \ref{fig:osm_goof}).  

As a means of comparison between OSM and SpaceNet labels, we use our baseline algorithm to train two models on SpaceNet imagery. One model uses ground truth masks rendered from OSM labels, while the other model uses the exact same algorithm, but uses ground truth segmentation masks rendered from SpaceNet labels.  Table \ref{tab:osm_vs_sn} displays APLS scores computed over
a subset of the SpaceNet
 test chips, and demonstrates that the model trained and tested on SpaceNet labels is far superior to other combinations, with a $\approx 60 - 90\%$ improvement.  This is likely due in part to the the more uniform labeling schema and validation procedures adopted by the SpaceNet labeling team.  

\begin{table}[h]
  \caption{OSM vs SpaceNet Performance}
  \label{tab:osm_vs_sn}
  \small
  \centering
   \begin{tabular}{llllll}
    \toprule
     Model & Test Labels & APLS \\
    \toprule
    OSM & OSM & 0.47 \\
    OSM & SpaceNet & 0.46 \\
    SpaceNet & OSM & 0.39 \\
    SpaceNet & SpaceNet & 0.75 \\
     \bottomrule
  \end{tabular}
\end{table}

\section{Scaling to Large Images} \label{sec:cresi}

\begin{figure}[]
\begin{center}
\includegraphics[width=0.95\linewidth]{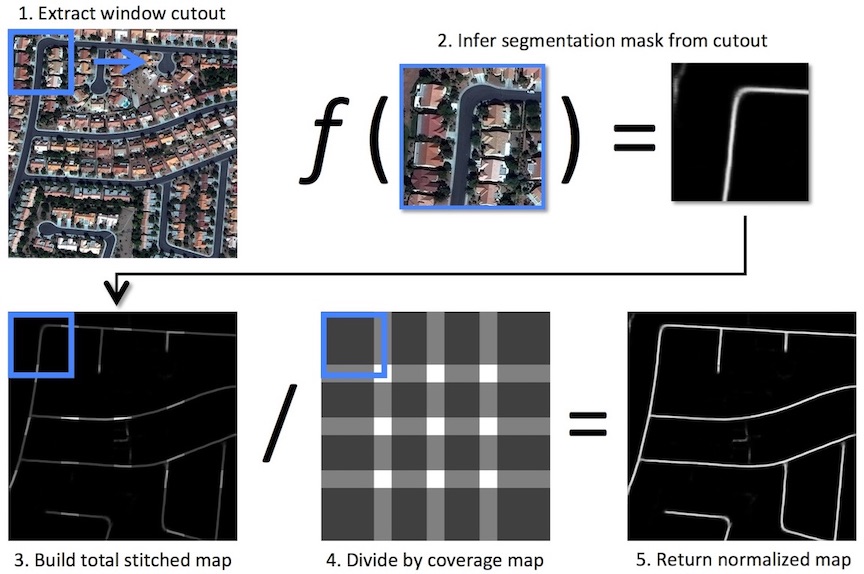}
\end{center}
\caption{Process of slicing a large satellite image (top) and ground truth road mask (bottom) into smaller cutouts for algorithm training or inference \cite{basiss}.}
\label{fig:SIMRDWN_training}
\vspace{-6pt}
\end{figure}

The process detailed in Section \ref{sec:algo} works well for small input images below $\sim2000$ pixels in extent, yet fails for images larger than this due to a saturation of GPU memory.  For example, even for a relatively simple architecture such as U-Net, typical GPU hardware (NVIDIA Titan X GPU with 12 GB memory) will saturate for images $> 2000$ pixels in extent and reasonable batch sizes $\leq 4$. 
In this section we describe a straightforward methodology for scaling up the algorithm to larger images.  
We call this approach 
City-scale Road Extraction from Satellite Imagery (CRESI).
The first step in this methodology provided by the Broad Area Satellite Imagery Semantic Segmentation (BASISS) \cite{basiss} methodology; this approach is outlined in Figure \ref{fig:SIMRDWN_training}, and returns a road pixel mask for a large test image.

We take this scaled road mask and apply Steps 3-6 of Table \ref{tab:algo} to retrieve the final road network prediction.  The final algorithm is given by Table \ref{tab:algo2}.  
The output of the CRESI algorithm is a {\it networkx} \cite{networkx} graph structure, with full access to the many algorithms included in the  {\it networkx} package.

\begin{table}[h]
  \caption{CRESI Inference Algorithm}
  \label{tab:algo2}
  \small
  \centering
   \begin{tabular}{ll}
    \toprule
     Step & Description \\
    \toprule
    	1 & Split large test image into smaller windows \\
	2 & Apply the 4 trained models to each window \\
	3 & For each window, merge the 4 predictions \\
	4 & Build the total normalized road mask \\
	5 & Clean road mask with opening, closing, smoothing \\ 
	6 & Skeletonize road mask \\ 
	7 & Extract graph from skeleton \\
	8 & Remove spurious edges and close small gaps in graph \\
    \bottomrule
  \end{tabular}
\end{table}

\section{Large Area Testing}

\subsection{Test Data}\label{sec:testdata}

We extract test images from all four of the SpaceNet cities with road labels: Las Vegas, Khartoum, Paris, and Shanghai.  As the labeled SpaceNet regions are non-contiguous and irregularly shaped,
we define rectangular subregions of the images where labels do exist within the entirety of the region.  The SpaceNet road labels in these regions form our ground truth, while the images serve as input to our trained model.  The regions are defined below in Table \ref{tab:test_regs}. 

\begin{table}[h]
  \caption{Test Regions}
  \label{tab:test_regs}
  \small
  \centering
   \begin{tabular}{lll}
    \toprule
     Test Region & Area & Road Length \\
     & (Km$^2$) & (Total Km) \\
    \toprule
    	Khartoum$\_$0 	& 3.0 	& 76.7 \\
	Khartoum$\_$1 	& 8.0 	& 172.6 \\
	Khartoum$\_$2 	& 8.3 	& 128.9\\
	Khartoum$\_$3 	& 9.0 	& 144.4 \\
	Las Vegas$\_$0 	& 68.1	& 1023.9\\
	Las Vegas$\_$1	& 177.0	& 2832.8\\
	Las Vegas$\_$2	& 106.7 	& 1612.1\\
	Paris$\_$0 		& 15.8	& 179.9\\
	Paris$\_$1		& 7.5		& 65.4\\
	Paris$\_$2		& 2.2		& 25.9\\
	Shanghai$\_$0		& 54.6	& 922.1\\
	Shanghai$\_$1		& 89.8	& 1216.4\\
	Shanghai$\_$2		& 87.5	& 663.7\\
	\hline
	Total			& 608.0	& 9064.8\\
    \bottomrule
  \end{tabular}
\end{table}

\section{Results}\label{sec:results}

\begin{figure}
\vspace{-6pt}
\begin{center}
\includegraphics[width=0.94\linewidth]{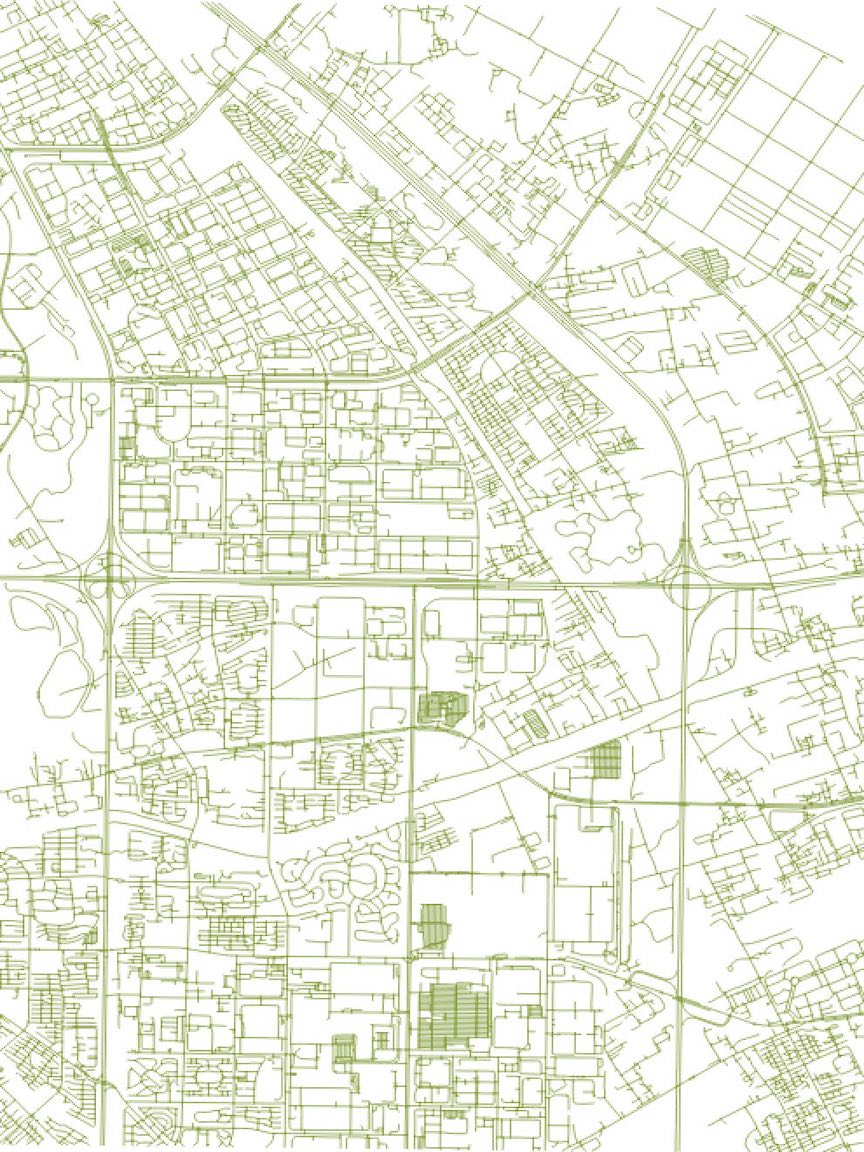}
\end{center}
\vspace{-6pt}
\caption{Output of CRESI inference as applied to the Shanghai$\_0$ test region. 
}
\label{fig:res_r0}
\vspace{-6pt}
\end{figure}

\begin{figure}
\begin{center}
\includegraphics[width=0.97\linewidth]{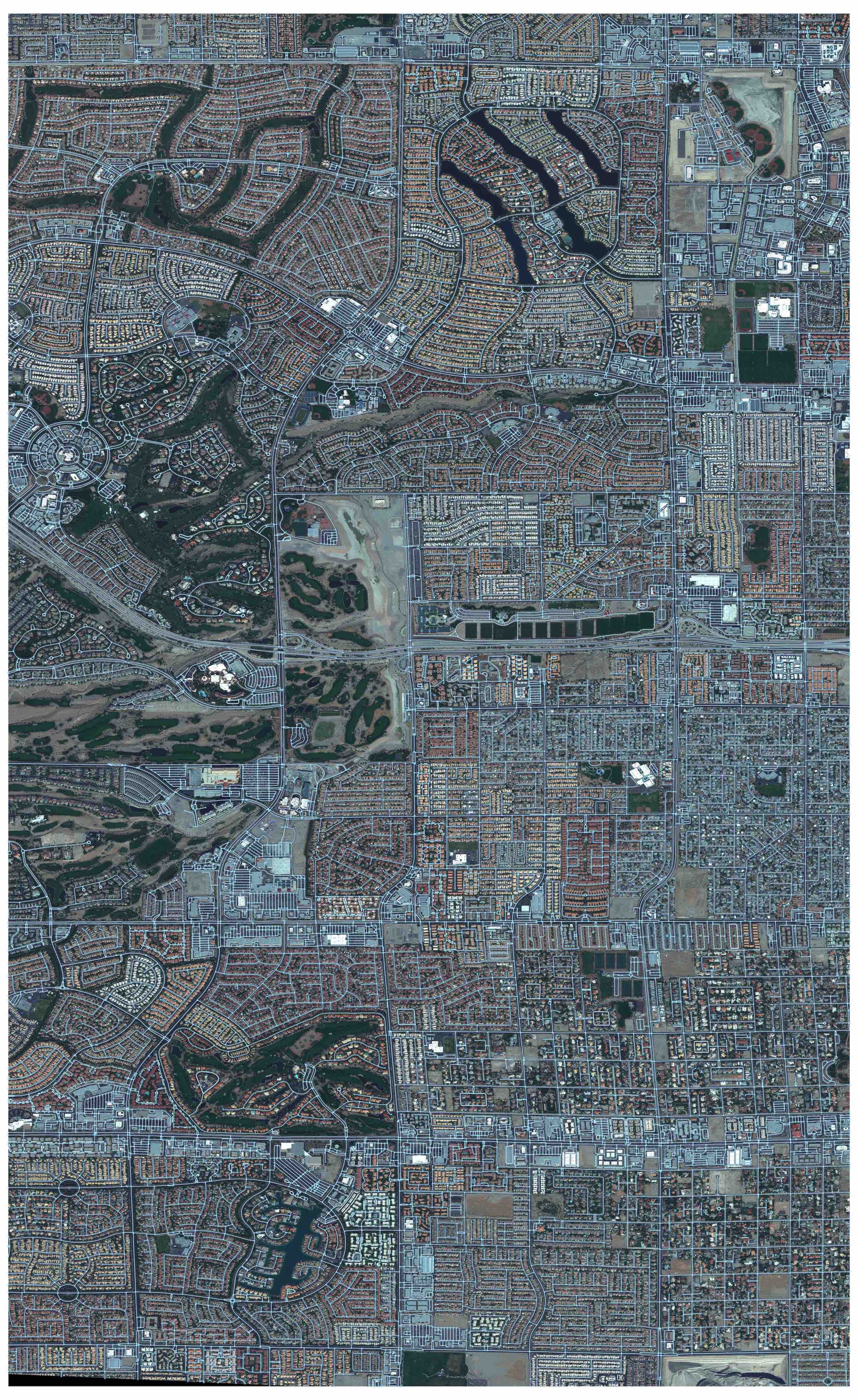}
\end{center}
\caption{Output of CRESI inference as applied to the Vegas$\_0$ test region. The APLS score for this prediction is 0.86.
There are 1023.9 km of labeled SpaceNet roads in this region, and 1029.8 km of predicted roads.}
\label{fig:res_r00}
\vspace{-6pt}
\end{figure}

\begin{figure}
\begin{center}
\includegraphics[width=0.99\linewidth]{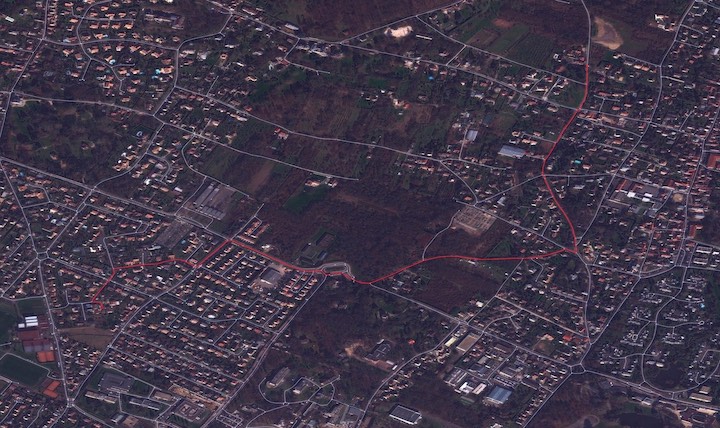}
\end{center}
\caption{Optimal route (red)  computed between two nodes of interest on the graph output of CRESI for a subset of the Paris$\_$0 test region.  
	}
\label{fig:res_r1}
\vspace{-6pt}
\end{figure}

We apply the CRESI algorithm described in Table \ref{tab:algo2} to the test images of Section \ref{sec:testdata}. Evaluation takes place with the APLS metric adapted for large images (no midpoints along edges and a maximum of 500 random control nodes), along with the TOPO metric, using an aggressive buffer size (for APLS) or hole size (for TOPO) of 4 meters.  
We report scores in Table \ref{tab:test_perf} as the weighted (by road length) mean and standard deviation of the test regions of Table \ref{tab:algo2}.  Figure \ref{fig:res_r0} displays the graph output of the algorithm.
Since the algorithm output is a \rm{ networkx} graph structure, myriad graph algorithms can by easily applied.  In addition, since we retain geographic information throughout the graph creation process, we can overlay the graph nodes and edges on the original GeoTIFF that we input into our model.  Figures \ref{fig:res_r00} and \ref{fig:res_r1} display portion of Las Vegas and Paris, respectively, overlaid with the inferred road network.  Figure \ref{fig:res_r1}demonstrates that road network extraction is possible even for atypical lighting conditions and off-nadir observation angles, and also that CRESI lends itself to optimal routing in complex road systems.


\begin{table}[h]
\vspace{-6pt}
  \caption{CRESI Performance}
  \label{tab:test_perf}
  \small
  \centering
   \begin{tabular}{lll}
    \toprule
     Test Region & APLS & TOPO \\
    \toprule
    	Khartoum  	& $ 0.64 \pm 0.07$ 	& $0.54 \pm 0.08$ \\
    	Las Vegas 	& $ 0.83 \pm 0.02$ 	& $0.65 \pm 0.01$ \\
    	Paris		 	& $ 0.64 \pm 0.03$ 	& $0.44 \pm 0.01$ \\
    	Shanghai  	& $ 0.51 \pm 0.01$ 	& $0.43 \pm 0.02$ \\
	\hline
	Total 		& $ 0.73 \pm 0.12$   & $0.58 \pm 0.09$ \\

    \bottomrule
  \end{tabular}
  \vspace{-6pt}
\end{table}

\subsection{Comparison to Previous Work}

Solutions from the recent SpaceNet 3 challenge maxed out at 0.67 \cite{sn3_solutions}.  The area-weighted total score of Table \ref{tab:test_perf} corresponds to a score 0f 0.66 if using city-wide averaging as in the SpaceNet 3 challenge.  Recall from Section \ref{sec:apls} that we use a slightly different formulation of APLS than the SpaceNet challenge that yields a score $\approx 3 - 6\%$ lower than the SpaceNet challenge; therefore our reported score of 0.66 would yield a SpaceNet score of $\approx 0.68 - 0.70$.

The total TOPO score of 0.58 compares favorably with the RoadTracer implementation, which reports an F1 score of $\approx 0.43$ 
for a larger (less restrictive) TOPO hole size, or a TOPO F1 score of $\approx 0.37$ for the DeepRoadMapper implementation (Figure 8 of \cite{roadtracer}).

The RoadTracer work used OSM labels and 0.6 m resolution aerial imagery.  To perform a more direct comparison to this work, we degrade imagery to 0.6 m resolution and train a new model; we also adopt a TOPO hole size of 12 meters to compare directly with the RoadTracer TOPO scores.  With the model trained (and tested) on 0.6 m data we observe a decrease of $14\%$ in the APLS score, to $0.63 \pm 0.22$.  The TOPO score actually rises slightly to $0.60 \pm 0.22$ due to the less stringent hole size.  This TOPO score represents a $40\%$ improvement over the RoadTracer implementation, though we caveat that testing and training is still on different cities for CRESI and RoadTracer, and SpaceNet labels were shown in Section \ref{sec:osm} to provide a significant improvement over OSM labels (which RoadTracer uses).  
A direct comparison between these methods is reserved for a later work.  

\subsection {Inference Speed}
Inference code has not been optimized for speed, but even so inference runs at a rate of $160 \,{\rm km}^2$ (approximately the area of Washington D.C.) per ${\rm hour}$  on a single GPU machine.  On a four GPU cluster the speed is a minimum of $370 \, {\rm km}^2 / \, {\rm hour}$.

\section{Conclusion}

Optimized routing is crucial to a number of challenges, from humanitarian to military. Satellite imagery may aid greatly in determining efficient routes, particularly in cases of natural disasters or other dynamic events where the high revisit rate of satellites may be able to provide updates far faster than terrestrial methods.

In this paper we demonstrated methods to extract city-scale road networks directly from remote sensing imagery.  GPU memory limitations constrain segmentation algorithms to inspect images of size $\sim1000$ pixels in extent, yet any eventual application of road inference must be able to process images far larger than a mere $\sim300$ meters in extent. Accordingly, we demonstrate methods to infer road networks for input images of arbitrary size. This is accomplished via a multi-step algorithm that segments small image chips, extracts a graph skeleton, refines nodes and edges, and stitches chipped predictions together to yield a final large road network graph.  Measuring performance with the APLS graph theoretic metric we observe superior performance for models trained and tested on SpaceNet data over OSM data.  Over four SpaceNet test cities, we achieve a total score of APLS = $0.73 \pm 0.12$, and inference speed of $160 \, {\rm km} ^2 \, / \, \rm{hour}$.  The TOPO scores are a significant improvement over existing methods, partly due to the higher fidelity labels provided by SpaceNet over OSM.  High fidelity road networks are key to optimized routing and intelligent transportation systems, and this paper demonstrates one method for extracting such road networks in resource starved or dynamic environments.

\bibliographystyle{IEEEtran}
\bibliography{bib}


\end{document}